# Back analysis based on SOM-RST system

H. Owladeghaffari

*Department of mining & metallurgical engineering, Amirkabir university of technology, Tehran, Iran*

H. Aghababaei

*Faculty of mining engineering, Sahand university of technology, Tabriz, Iran*

ABSTRACT: This paper describes application of information granulation theory, on the back analysis of "Jeffrey mine- southeast wall-Quebec". In this manner, using a combining of Self Organizing Map (SOM) and rough set theory (RST), crisp and rough granules are obtained. Balancing of crisp granules and sub rough granules is rendered in close-open iteration. Combining of hard and soft computing, namely finite difference method (FDM) and computational intelligence and taking in to account missing information are two main benefits of the proposed method. As a practical example, reverse analysis on the failure of the southeast wall-Jeffrey mine – is accomplished.

## 1 INTRODUCTION

Back analysis is a reverse procedure, which is to solve the external load or partial material parameters, based on the known deformation and stresses at limited points and the partially known material parameters (Zhu&Zhao, 2004).

In most geotechnical engineering problems, it is often necessary to know the in situ field, material mechanical parameters, and even the mechanical model, by utilizing the monitored physical information such as deformation, strain, stress, and pressure during construction. The back analysis method, based on the required input physical deformation, can be divided in to deformation back analysis method, stress back analysis method and coupled back analysis method (Sakurai et al, 2003). The complex feature of rock mass, associated structures and the difficulty of interpretation of the interaction of excavation and rock mass are the main reasons to deploying of back analysis methods in rock engineering. Generally, two main procedures of back analysis methods have been pointed in the literature: inverse and direct approach (Cividini et al, 1989). In reverse method, the mathematical formalization is inverse of ordinary analysis. In this process, the number of monitored data is more than the unknown parameters, so that by utilizing of optimization techniques, the unknown parameters are obtained. The main advantage of such method is an independent spirit in regard of the iteration operations, which causes to reducing of calculation time, sensibly. Direct approach, is a process associated with the cycling optimization (aim is to minimize the fitness function). Application of direct approach, to evaluation of problem based non-linearity and employing of heuristic optimization methods such Ant Colony (ACO), Particle Swarm (PSO), and Simulated Annealing (SA) are two main results. So, back analysis based on interactions of hard computing and soft computing methods can be settled in direct approach. As an initial over view, 1-1 mapping methods can be supposed as hard computing such analytical, numerical or hybrid methods, whereas soft computing methods (SC), as main part of not 1-1 mapping methods, is a coalition of methodologies which are tolerant of imprecision, uncertainty and partial truth when such tolerance serves to achieve better performance, higher autonomy, greater tractability, lower solution cost. The principles members of the coalition are fuzzy logic (FL), neuro computing (NC), evolutionary computing (EC), probabilistic computing (PC), chaotic computing (CC), and machine learning (Zadeh, 1994).

Utilizing of soft computing methods besides hard computing methods, such support vector machine (Feng et al, 2004) and neural network (Pichler et al, 2003), in the reverse procedures have been experienced, successfully. Advancing of soft computing methods, under the general new approach, namely information granulation theory, has been opened new horizons on the knowledge discovery data bases. Information granules are collections of entities that are arranged due to their similarity, functional adjacency, or indiscernibility relation. The process of forming information granules is referred to as information granulation (Zadeh, 1997). There are many approaches to construction of IG, for example Self Organizing Map (SOM) network, Fuzzy c-means (FCM), rough set theory (RST). The granulation level, depend on the requirements of the project. The smaller IGs come from more detailed processing. On the other hand, because of complex innate feature of information in real world and to deal with vagueness, adopting of fuzzy and rough analysis or the combination form of them is necessary. In fact, because of being the complex innate of crisp, rough or fuzzy attributes, in the natural world, extraction of fuzzy or rough information inside the crisp granules, can give a detailed granulation. To develop back analysis based new soft computing approaches and taking in to account of a mathematical tool for the analysis of a vague description of objects, as well as rough set theory, a combining of self organizing map-neural network-(SOM), and rough set theory(RST) associated with the hard computing methods is proposed.

The rough set theory introduced by Pawlak (1982, 1991) has often proved to be an excellent mathematical tool for the analysis of a vague description of object. The adjective vague, referring to the quality of information, means inconsistency, or ambiguity which follows from information granulation. The rough set philosophy is based on the assumption that with every object of the universe there is associated a certain amount of information, expressed by means of some attributes used for object description. The indiscernibility relation (similarity), which is a mathematical basis of the rough set theory, induces a partition of the universe in to blocks of indiscernible objects, called elementary sets, which can be used to build knowledge about a real or abstract world. Precise condition rules can be extracted from a discernibility matrix. The rest of paper has been organized as follow: a part from details, section 2 covers a brief on the RST and SOM and in section 3 the details of the proposed algorithm has been presented. Finally, we discuss an example in which we show how the proposed technique behaves.

## 2 RST&SOM

Here, we present some preliminaries of self organizing feature map –neural network- and rough set theory that are relevant to next section.

### 2.1 Self organizing map-neural network

Kohonen's (1987) SOM algorithm has been well renowned as an ideal candidate for classifying input data in an unsupervised learning way.

Figure 1 shows Kohonen's SOM topology, where two layers such as input layer and mapping layer are working together to organize input data into an appropriate number of clusters (or groups).

For supervised learning, both input and output are necessary for training the neural network, while the unsupervised learning needs only the inputs. In unsupervised learning, neural network models adjust their weights so that input data can be organized in accordance with statistical properties embedded in input data. Kohonen's SOM includes two layers such as input layer and mapping layer, in the shape of a one or two-dimensional grid. The number of nodes in the input layer is equal to the number of features associated with input data. The mapping (output) layer acts as a distribution layer. Each node of the mapping layer or output node also has the same number of features as there are input nodes. Both layers are fully connected and each connection is given an adjustable weight. Furthermore, each output node of the mapping layer is restricted smaller distance around the cluster center.

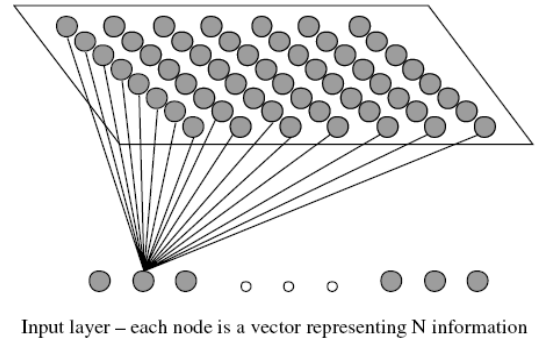

Figure 1. Kohonen's SOM

For a Kohonen's SOM, suppose that the number of output nodes is m, the number of input nodes is $n$, and $w_i = (w_{i1}, w_{i2}, ... w_{im})(1 \leq i \leq m)$ is the connection weight vector corresponding to output node $i$. $w_i$ can be viewed as a center of the cluster $i$. whenever new input data $x = (x_1, x_2, ..., x_n)$ is presented to SOM during the training phase, the output value for output node $i$ is computed by square of the Euclidean distance denoted by $o_i$ between $x$ and $w_i$, as shown in Equation 1:

$$o_i = (d_i)^2 = \|x - w_j\|^2 = \sum_{j=2}^{n}(x_j - w_{ij})^2, 1 \leq i \leq m \quad (1)$$

If the node $i^*$ satisfies Equation 2 then it is declared as a winner node:

$$(d_{i^*})^2 = \min o_i, 1 \leq i \leq m \quad (2)$$

Adjustable output nodes including the winning node $i^*$ and its neighbor nodes are determined by the neighborhood size of the winning node $i^*$ ($|\Omega_{i^*}|$). Subsequently, connection weights of the adjustable nodes are all updated. The learning rule of SOM is shown in Equation 3:

$$\Delta w_{ij} = \eta(x_j - w_{ij}), i \in |\Omega_{i^*}|, 1 \leq j \leq n \quad (3)$$

Where $\eta$ is a learning rate. To achieve a better convergence, $\eta$ and the neighborhood size of the winning node, should be decreased gradually with learning time.

To achieve a better convergence, and the neighborhood size of the winning node, should be decreased gradually with learning time. SOM has been successfully employed in different fields of applied science. Specially, in geomechanics, for example, in clustering of lugeon data (Shahriar&Owladeghaffari, 2007) and joint sets (Sirat& Talbot, 2001).

### 2.2 Rough set theory

The rough set theory introduced by Pawlak (Pawlak,

1991) has often proved to be an excellent mathematical tool for the analysis of a vague description of object. The adjective vague referring to the quality of information means inconsistency, or ambiguity which follows from information granulation. The rough set philosophy is based on the assumption that with every object of the universe there is associated a certain amount of information, expressed by means of some attributes used for object description. The indiscernibility relation (similarity), which is a mathematical basis of the rough set theory, induces a partition of the universe in to blocks of indiscernible objects, called elementary sets, which can be used to build knowledge about a real or abstract world. Precise condition rules can be extracted from a discernibility matrix.

An information system is a pair $S=<U, A>$, where U is a nonempty finite set called the universe and A is a nonempty finite set of attributes. An attribute a can be regarded as a function from the domain U to some value set $V_a$. An information system can be represented as an attribute-value table, in which rows are labeled by objects of the universe and columns by attributes. With every subset of attributes $B \subseteq A$, one can easily associate an equivalence relation $I_B$ on U:

$$I_B = \{(x,y) \in U : for\ very\ a \in B, a(x) = a(y)\}$$

Then, $I_B = \bigcap_{a \in B} I_a$.

If $X \subseteq U$, the sets $\{x \in U : [x]_B \subseteq X\}$ and $\{x \in U : [x]_B \cap X \neq \varphi\}$, where $[x]_B$ denotes the equivalence class of the object $x \in U$ relative to $I_B$, are called the B-lower and the B-upper approximation of X in S and denoted by $\underline{BX}$ and $\overline{BX}$, respectively. It may be observed that $\underline{BX}$ is the greatest B-definable set contained in X and $\overline{BX}$ is the smallest B-definable set containing X To constitute reducts of the system, it will be necessary to obtain irreducible but essential parts of the knowledge encoded by the given information system. So one is, in effect, looking for the maximal sets of attributes taken from the initial set $(A, say)$ that induce the same partition on the domain as A. In other words, the essence of the information remains intact, and superfluous attributes are removed. Consider $U=\{x1, x2, …, xn\}$ and $A=\{a1, a2, …, an\}$ in the information system $S=<U, A>$. By the discernibility matrix M(S) of S is meant an n*n matrix such that $c_{ij} = \{a \in A : a(x_i) \neq a(x_j)\}$

A discernibilty function fs is a function of m Boolean variables a1,…,am corresponding to attributes a1,…,am ,respectively, and defined as follows:

$$f_s(a_1,...,a_m) = \wedge\{\vee(c_{ij}) : 1 \leq i,j \leq n, j \prec i, c_{ij} \neq \varphi\}$$

where $\vee(c_{ij})$ is the disjunction of all variables with a. $a \in c_{ij}$ (Pal et al, 2004). With such discriminant matrix the appropriate rules are elicited.

The existing induction algorithms use one of the following strategies:

(a) Generation of a minimal set of rules covering all objects from a decision table;
(b) Generation of an exhaustive set of rules consisting of all possible rules for a decision table;
(c) Generation of a set of `strong' decision rules, even partly discriminant, covering relatively many objects each but not necessarily all objects from the decision table (Greco et al, 2001). In this study we use first approach in combining with initial crisp granules, adaptively.

## 3 THE PROPOSED METHOD BASED ON SOM-RST SYSTEM AND HARD COMPUTING METHODS

Figure 2, shows a general procedure, in which the information granulation theory accompanies by a predefined project based rock engineering design. After determining of constraints and the associated rock engineering regards, the initial granulation of information as well as numerical (data base) or linguistic format is accomplished. Developing of modeling instruments based on IGs, whether in independent or associated shape with hard computing methods (such fuzzy finite element, fuzzy boundary element, stochastic finite element…) are new challenges in the current discussion.

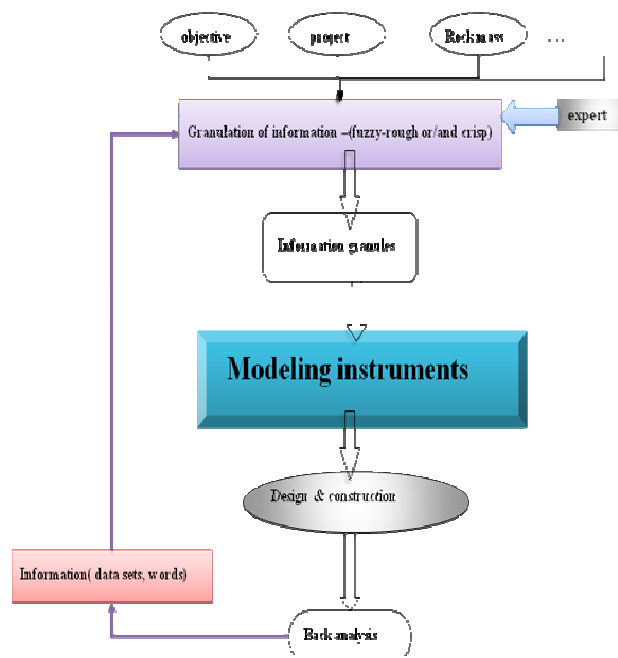

Figure 2. A general methodology for back analysis based on information granulation theory.

Thus, one can employ such method as a new mythology in designing of rock engineering flowcharts. The main benefits are considering of the roles of the expert's experiences and educations, the missing or vagueness information and utilizing of advantages of soft and hard computing methods. A new advantages of the mentioned method, is to considering of the outward changes in *"words format"* and *"calculation with words and perception" (CWP)*.

Under mentioned methodology and to develop modeling instruments, authors have (are) proposed (proposing) new algorithms with neural networks, fuzzy logic (possibility theory), rough set theory, and meta heuristic optimization methods which are accompanied by close-open world idea. Random selection of initial precise granules can be set as "close world" assumption (CWA). But in many applications, the assumption of complete information is not feasible (CWA) nor realistic, and only cannot be used. In such cases, an open world assumption (OWA), where information not known by an agent is assumed to be unknown, is often accepted. The aim of open-world is to achieve complete knowledge of the universe by set of classified rules or by modifying rules. The close-open iteration accomplishes the balancing of crisp and sub rough granules by some random selection of initial granules and increment (or decreasing) of supporting rules, gradually. Figure 3, shows one of our proposed algorithms, apart from contributing of FL and free-derivative optimization methods. One may employ other shape of open world assumption's implementation. For instance to raise the quality of approximation by much more categories, but the optimum numbers of such scaling could be approached by an algorithm, so that an adaptive disceritization besides balancing of the initial granules with open world is accomplished

First step is to collect data sets by using 1-1 modeling (here, hard computing methods as one of the main part of the direct modeling methods are selected). Applying of SOM as a preprocessing step and discretization tool is second process. For continuous valued attributes, the feature space needs to be discretized for defining indiscernibilty relations and equivalence classes. We discretize each feature in to three levels by SOM: *"low, medium, and high"*; finer discretization may lead to better accuracy at the cost of a higher computational load. Because of the generated rules by a rough set are coarse and therefore need to be fine-tuned, here, we have used the preprocessing step on data set to crisp granulation by SOM, so that extraction of best initial granules and then rough granules is rendered in close-open iteration. In open world iteration phase, we use a simple idea associated with human's cognition of the surround world:*"simplicity of rules"* (dominant to the problem) whether in numbers, applied operators or/and the length of rules. Here, we take in to account three main parameters in rules generation: length of rules, strength of object and number of rules. By setting the adaptive threshold error level and the number of close-open iteration, stability of the algorithm is guaranteed. After obtain best rules, the monitored data set as real decision parts are compared with the extracted rules and best conditional parts are picked up. These parts, associated with the reduced data sets by SOM, are the approximation of the wanted parameters as well as back analysis results.

## 4 AN ILLUSTRATIVE EXAMPLE: FAILURE ON THE SOUTHEAST WALL- JEFFREY MINE

This section describes how one can acquired approximated values of the effective parameters under a reverse analysis. To this aim and by using finite difference method, as well as FLAC (), the failure on the southeast wall in Jeffrey mine (Quebec) has been evaluated.

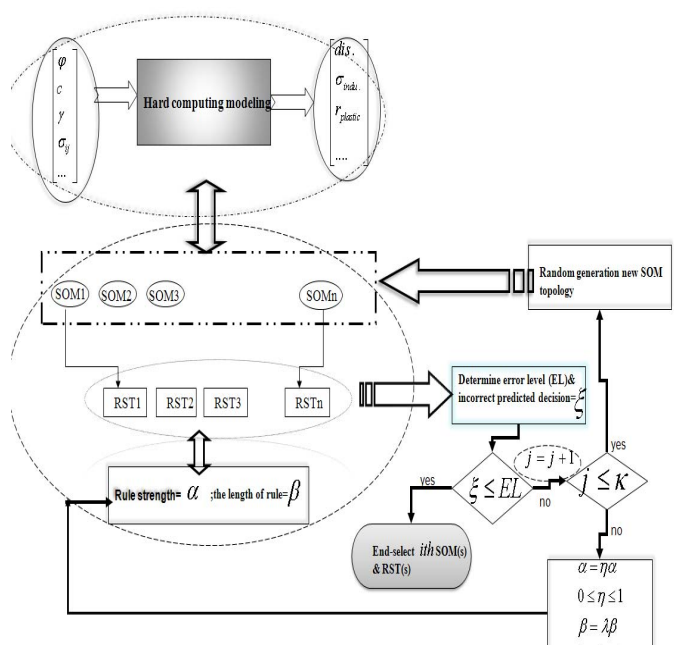

Figure 3. A proposed procedure based on hard computing and SOM-RST system.

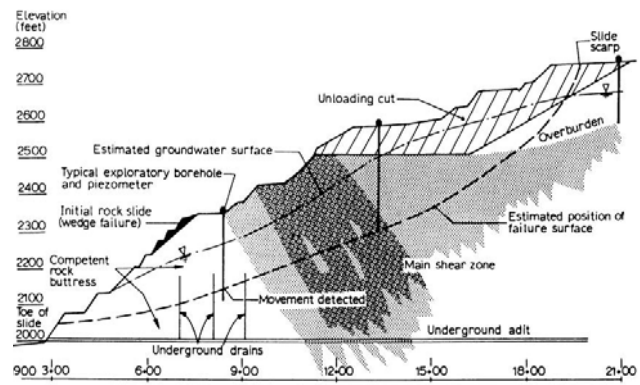

Figure 4. Cross section through the Jeffry mine showing the 1971 failure surface (Sjoberg, 1996)

| NO | Cp*10^5 | Phi p | Cb | Phi b | C sz | Phi sz | T p | T b | TMD | MVV |
|----|---------|-------|----|----|------|--------|-----|-----|------|-----|
| 1 | 2.00 | 25 | 3.00E+05 | 35 | 500 | 15 | 428901.4 | 42844.4 | 2.41E-05 | 4.46E-14 |
| 2 | 3.20 | 35 | 2.20E+05 | 25 | 1000 | 5 | 428901.4 | 42844.4 | 5.82E-06 | 9.86E-16 |
| 3 | 2.50 | 30 | 3.50E+05 | 40 | 1500 | 10 | 428901.4 | 42844.4 | 2.66E-06 | 1.84E-14 |
| 4 | 2.75 | 25 | 3.75E+05 | 35 | 1500 | 15 | 428901.4 | 42844.4 | 4.27E-06 | 1.68E-22 |
| 5 | 3.00 | 35 | 4.00E+05 | 45 | 1000 | 20 | 428901.4 | 42844.4 | 2.52E-08 | 2.06E-22 |
| 6 | 2.00 | 25 | 3.00E+05 | 35 | 1500 | 15 | 428901.4 | 42844.4 | 2.61E-06 | 4.24E-14 |
| 7 | 2.00 | 25 | 3.00E+05 | 35 | 1500 | 15 | 1.00E+06 | 6.80E+05 | 2.20E-11 | 4.10E-22 |
| 8 | 2.00 | 25 | 3.00E+05 | 35 | 1500 | 5 | 1.00E+06 | 6.80E+05 | 2.58E-06 | 1.73E-13 |
| 9 | 2.00 | 25 | 3.00E+05 | 35 | 500 | 15 | 2.71E+06 | 1.13E+06 | 7.80E-16 | 1.92E-22 |
| 10 | 2.00 | 25 | 3.00E+05 | 35 | 800 | 7 | 2.71E+06 | 1.13E+06 | 3.71E-08 | 4.95E-13 |
| 11 | 2.00 | 25 | 3.00E+05 | 35 | 800 | 8 | 2.71E+06 | 1.13E+06 | 1.76E-07 | 1.70E-15 |
| 12 | 2.00 | 25 | 3.00E+05 | 35 | 500 | 8 | 2.71E+06 | 1.13E+06 | 1.16E-06 | 8.14E-16 |

Table1. Subset of the obtained results by FLAC4- Cp (kg/cm^2): coherent of Periditotets; Sz: shear zone;T: tensile strength(kg/cm^2); TMD: total max. Displacement (m); MVV: max. Velocity vector (m/s)

Asbestos fiber is mined from ultra basic host rocks dominated by periditotes, dunites and serpantites. The rocks mass is intersected by several thick shear zones and smaller scale discontinuous. Strength and deformability for the rock materials vary widely from very soft and weak to moderato stiff and strong rock. In 1970, the slope height was 180 m plus 60 m of overburden (clay, silt, &sand). Failure occurred in relatively fractured, serpantinized peridotite with a major shear zone present. Sliding in the overburden was first observed, followed by local wedge failure and finally a major slide in 1971, involving some 33 million ton of rock (figure4). Failure was believed to have standard in the weak shear zone which then leads to failure of upper portions of the slope movement rates of the order of 1500 $m/month$ were recorded. After this failure, some reactivation of the 1971 of the 1971 year failure occurred (Sjoberg, 1996).

By using finite difference method under *FLAC4*, as a hard computing method, almost 30 different models on the mentioned slope and by considering of Mohr-Cloumb plastic law have been constructed. The aim was to detect the material properties, while the slope movement rate satisfies the monitored records, approximately. Table 1shows the part of the obtained results while 8 effective parameters changes.

The performance of $3*3$ SOM ($n_x=3, n_y=3$; Matrix of neurons - $n_x \cdot n_y$ determines the size of 2D SOM.) on the maximum velocity vectors of the tested data have been depicted in figures 5 and 6 (The discritization procedure using $3*1$ SOM on the attribute has been rendered). This step by accounting of different structures of SOM and in interaction with RST was iterated. The effective rough set procedure parameters were selected as follows: $minimum\ rule\ strength: 60\%;$ *maximum length of rule: 2 and maximum number of rules: 5.* It must be noticed that decreasing of the rules is accomplished, gradually (depend on the error level).

Error level, here, is settled in true classified test data (percentage). Since total of analysis is low (30), we set *n=1, EL=80% and* $r=2$ *(maximum closed-iterations for any incensement of rules).*

After four closed-open iterations, algorithm satisfies conditions of the test data. Figure 7 illustrates the extracted rules which are matched with the recorded data by the proposed algorithm.

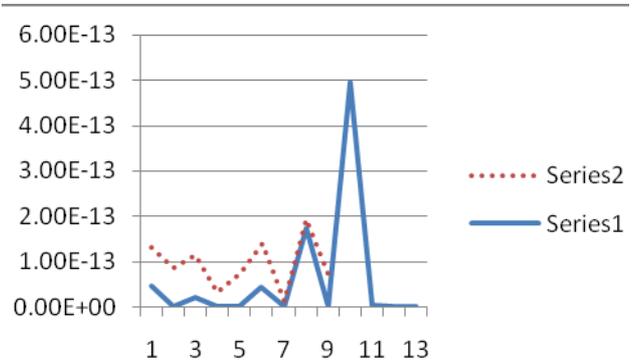

Figure 5. The performance of $3*3$ SOM on the maximum velocity vectors of the tested data (series 1: real; series 2: deduced data set)

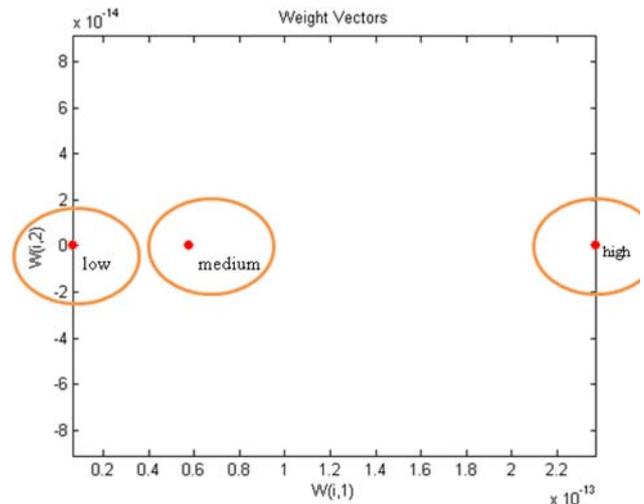

Figure 6. Discritization on the TMV in three clusters: low (3), medium (2), and high (1) by SOM.

So, we could extract some best rules where the decision parts of them satisfy the monitoring rate of the slope. Thus, Application of such system under regular monitoring data set can be rendered. The main point in such data is to granulate the real deci-sions and compare with the deduced decision parts rules.

```
Rule 1.  (cb<=220000.000000)  => (mvv at most 1);
Rule 2.  (phib<=25.035000)  => (mvv at most 1);
Rule 3.  (csz<=999.790000)  => (mvv at most 1);
Rule 4.  (phisz<=5.035400)  & (tb<=42844.000000) => (mvv at most 1);
```

Figure 7. Part of the extracted rules by the proposed algorithm

## 5 CONCLUSIONS

The role of uncertainty and vague information in geomechnaic analysis is undeniable feature. Indeed, with developing of new approaches in information theory and computational intelligence, as well as soft computing approaches, it is necessary to consider these approaches within and inside of current and conventional analysis, especially in geomechanic field. Under this idea and to fining best information granules which are picked off inside each other, close-open worlds (cycle) procedure has been proposed. Utilizing this algorithm in interaction with finite difference method and inferring back analysis results are main advantages of our method. So, we could obtain a becoming approximation of the parameters (in failure of southeast wall of Jeffry mine -1971). It must be notice that application of such process ensues a (or some) reduct set of the attributes where the reduct set(s) can be supposed as soft sensitive analysis on the model. Decreasing of time consuming and extractions of effective parameters, behind the core(s)-most effective parameters- of attributes (in condition parts) are other benefits of our model.